\pdfoutput=1

\documentclass[11pt]{article}

\usepackage{naacl2021}

\usepackage{times}
\usepackage{latexsym}

\usepackage[T1]{fontenc}

\usepackage[utf8]{inputenc}

\usepackage{microtype}
\usepackage{multirow}
\usepackage{booktabs}
\usepackage{comment}
\usepackage{array}
\usepackage{graphicx}
\graphicspath{ {./figures/} }
\usepackage{latexsym}
\usepackage{amsmath}
\usepackage{enumitem}

\title{Benchmarking Commercial Intent Detection Services with Practice-Driven Evaluations}

\author{Haode Qi$\dagger^*$, Lin Pan$\dagger$\thanks{$^*$Equal contribution}, Atin Sood$\dagger$, Abhishek Shah$\dagger$ \\ {\bf Ladislav Kunc$\dagger$, Mo Yu$\ddagger$, Saloni Potdar$\dagger$} \\
  $\dagger$IBM Watson \\
  $\ddagger$MIT-IBM Watson AI Lab \\
  {\small\tt \{Haode.Qi,Abhishek.Shah1,lada\}@ibm.com} \\ {\small\tt\{panl,asood,yum,potdars\}@us.ibm.com}}
  
\begin{document}
\maketitle
\begin{abstract}

Intent detection is a key component of modern goal-oriented dialog systems that accomplish a user task by predicting the intent of users' text input.
There are three primary challenges in designing robust and accurate intent detection models.
First, typical intent detection models require a large amount of labeled data to achieve high accuracy. Unfortunately, in practical scenarios it is more common to find small, unbalanced, and noisy datasets.
Secondly, even with large training data, the intent detection models can see a different distribution of test data when being deployed in the real world, leading to poor accuracy.
Finally, a practical intent detection model must be computationally efficient in both training and single query inference so that it can be used continuously and re-trained frequently.
We benchmark intent detection methods on a variety of datasets.
Our results show that \mbox{Watson Assistant's} intent detection model outperforms other commercial solutions and is comparable to large pretrained language models while requiring only a fraction of computational resources and training data.
\mbox{Watson Assistant} demonstrates a higher degree of robustness when the training and test distributions differ.
\end{abstract}

\section{Introduction}

Intent detection and entity recognition form the basis of the Natural Language Understanding (NLU) components of a task-oriented dialog system. The intents and entities identified in a given user utterance help trigger the appropriate conditions defined in a dialog tree which guides the user through a predetermined dialog-flow. 
These task-oriented dialog systems have gained popularity for designing applications around customer support, personal assistants, and opinion mining, etc.

The Conversational AI market is expected to grow to an estimated USD 13.9 billion by 2025 as reported by Markets \& Markets \footnote{{https://customerthink.com/conversational-ai-in-2021-3-top-trends-to-look-out-for}}. There are several solutions in the market that help enterprises build and deploy chatbots quickly to automate large portions of their customer service interactions. Hence, a commercial conversational AI solution needs to adapt to a variety of use cases, accurately identify users' intents and resolve their queries. 

\begin{figure}[t]
\includegraphics[width=\linewidth]{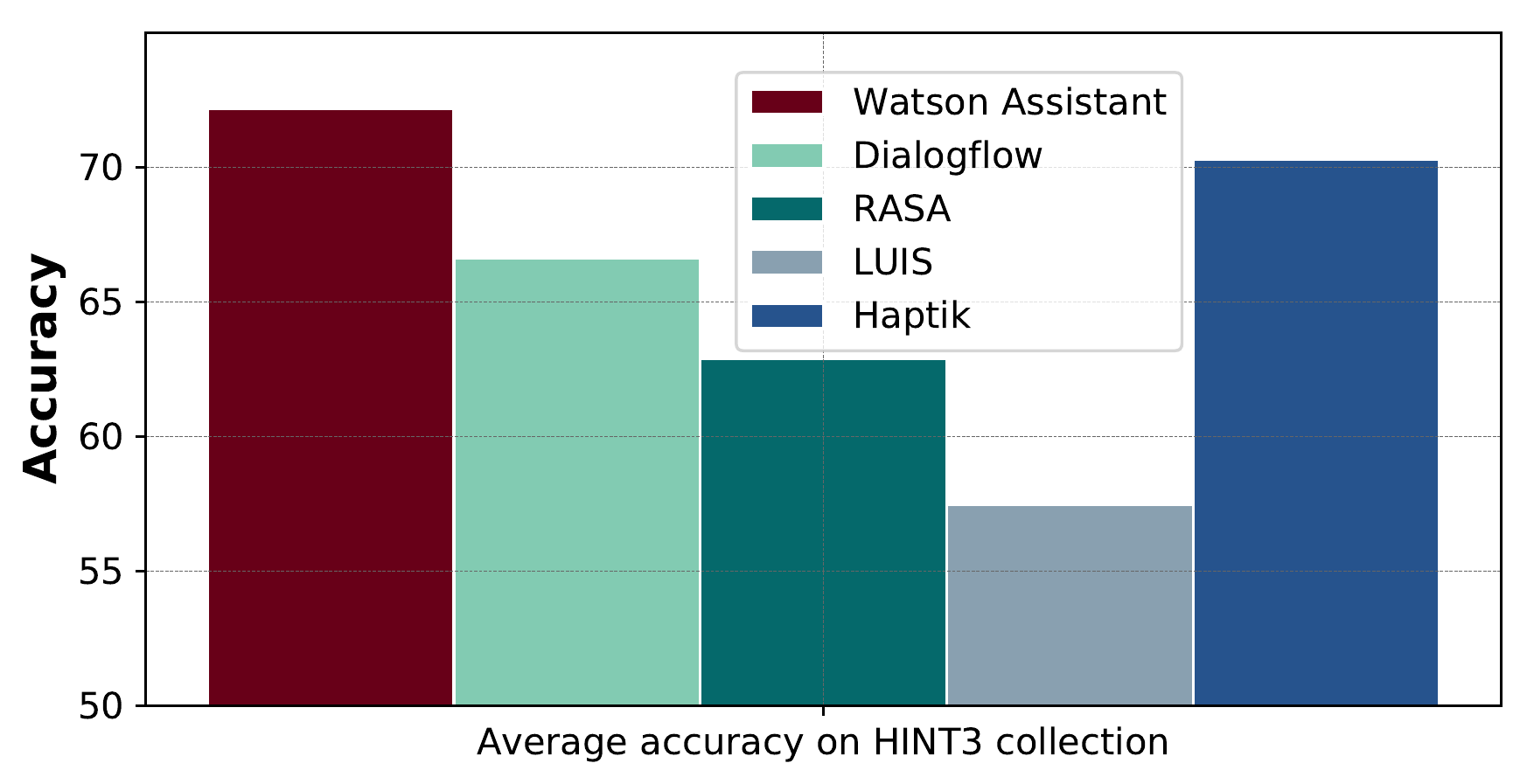}
\caption{\textbf{Accuracy of commercial solutions} on the HINT3 collection of datasets. Results are averaged across the \emph{Full} versions of the three datasets and their \emph{Subset} versions. The in-scope accuracy is reported on a threshold of 0.1. Watson Assistant achieves the best results on average. Results for all methods except Watson Assistant are obtained from \citet{Arora2020HINT3RT}.}
\label{overall-market}
\end{figure}

There are three primary challenges in designing intent detection models that power real-world dialog systems: (1) \textbf{Limitations in training data}: while typical machine learning models are trained on large, balanced, labeled datasets, practical intent detection systems rely on customer provided data. These datasets are usually small, probably noisy, unbalanced, and contain classes with overlapping semantics, etc. The relatively poor quality of training data makes it hard to train accurate models.
(2) \textbf{Robustness to non-standard user inputs}: when the intent detection models are deployed in real-world settings, they often operate on test data that differs significantly from the training data. The mismatch in train and test data distributions mainly comes from the free-form of input user queries. These real world queries express the same intents with their non-standard paraphrases, which are difficult to fully cover during training. 
The lack of large and clean training data makes this problem worse.
(3) \textbf{Computational efficiency}: the intent detection models should be computationally efficient for both training and inference. On one hand, efficient inference is crucial since it allows for faster query resolution times for the users.\footnote{Inference time is usually dependent on service-level agreements between the provider and the user which determine the response time upper bounds of the APIs. This is hard to measure and compare across services in a reliable way for the purpose of this study.} On the other hand, a real-world dialog system is frequently updated according to customer needs, so faster training time becomes an important consideration for real-world conversational AI solutions. 




In this work, taking the aforementioned three realistic challenges into consideration, we evaluate multiple intent detection models and focus on their accuracy, data efficiency, robustness, and computational efficiency.
We compare the performance of various commercial intent detection models on three datasets in the HINT3 collection \cite{Arora2020HINT3RT}. 
We also evaluate pretrained Language Models (LM) on three commonly used public datasets for benchmarking intent detection - CLINC150 \cite{Larson+19}, BANKING77 \cite{Casanueva+20}, and HUW64 \cite{Liu+19}. 
In addition, we create few-shot learning settings from these datasets, to better match real world low-resource scenarios.
Furthermore, we measure the "in the wild"  robustness of the systems via creating difficult test subsets from existing test sets.
Finally, we evaluate the classification accuracy and training time of these models because it directly affects the usability and development life-cycle of an conversational AI solution.

We build upon the existing study in \citet{Arora2020HINT3RT} which benchmarked commercial solutions aside from IBM Watson Assistant (i.e.,  Dialogflow, LUIS, and RASA). We extend this study by adding Watson Assistant and recent large-scale pretrained LMs. We also explore few-shot and robustness settings, and compare the resource efficiency and training times of different commercial solutions as well as pretrained LMs. Among these solutions, Watson Assistant’s new intent detection algorithm performs better than other commercial solutions (Figure~\ref{overall-market}), and achieves comparable accuracy when compared to large-scale pretrained LMs (Figure~\ref{scatter-clinc}) while being much more efficient. 


\section{Related Work}

Several datasets have been released to test the performance of intent detection for task-oriented dialog systems such as Web Apps, Ask Ubuntu and Chatbot corpus from \citet{braun-etal-2017-evaluating}; ATIS dataset \cite{price1990evaluation} and SNIPS \cite{coucke2018snips}. The ATIS and SNIPS datasets have been created with focus on voice interactive chatbots. Voice modality has some specific characters, i.e., it does not contain typos and it is less noisy than text-based communication. Thus, these datasets are oversimplified version of the text-based intent detection task "in the wild" due to well-constructed dataset and limited number of classes. 

Recently, CLINC150 \cite{Larson+19}, BANKING77 \cite{Casanueva+20}, and HWU64 \cite{Liu+19} have been used to benchmark the performance of intent detection systems. These datasets cover a large number of intents across a wider range of domains, which captures more real-world complexity of doing fine-grained classification. \citet{Arora2020HINT3RT} proposed a new collection of datasets called HINT3, containing a noisy and diverse set of intents and examples across three domains sourced from domain experts and real users. 

Prior work from \citet{Arora2020HINT3RT}, \citet{braun-etal-2017-evaluating}, and \citet{Liu+2019} study the performance of different conversational AI services using the datasets mentioned above.
\citet{Casanueva+20}, \citet{Larson+19}, \citet{Arora2020HINT3RT}, \citet{bunk2020diet} and others have benchmarked several state-of-the-art (SOTA) pretrained LMs such as BERT \cite{Devlin+18} on the aforementioned datasets.

We aim to standardize the benchmarking tests that need to be run while developing an industry scale intent detection system. The tests should cover a variety of real-world datasets, settings such as few-shot scenarios and testing on semantically dissimilar test examples. 
Additionally, the tests should also cover resource efficiency and training time - since they affect the overall deployment costs of a virtual assistant cloud service. A carefully chosen trade-off between accuracy and efficiency is the decision making factor in choosing the algorithm for the real-world intent detection system. 


\section{Evaluation Settings}
\label{datasets}
\subsection{Datasets}
We create our proposed evaluation settings based on the following public intent detection datasets:

\paragraph{CLINC150} consists of $22,500$ in-scope examples that cover $150$ intents in $10$ domains, such as banking, work, travel, etc. The dataset also comes with $1,200$ out-of-scope examples. In this work, we only focus on the in-scope examples. 

\paragraph{HWU64} contains $25,716$ examples, covering $64$ intents in $21$ domains. The data creation process aims to reflect human-home robot interaction. We are using one fold train-test split with $9,960$ training examples and $1,076$ testing examples.  

\paragraph{BANKING77} is a single domain dataset created for fine-grained intent detection. It focuses on the banking domain, and has $13,083$ examples covering $77$ intents. 

\subsection{Practice-Driven Benchmark Settings}

\paragraph{Full-set setting} This corresponds to the standard evaluation setting that uses the full training and testing sets.

\paragraph{Few-shot setting}
In the real-world setting, users may not provide a large number of labelled examples to train a conversational AI system. Labeling data is extremely time consuming and difficult, so we need to make our intent detection systems robust enough to handle the few-shot scenarios and improve time to value for the user. 
We create a few shot setup for all the datasets by sampling 5 examples per intent and 30 examples per intent on CLINC150, HWU64 and BANKING77 datasets.

\paragraph{Difficult test setting}
Most of the current SOTA classification models can achieve $90\%$+ test accuracy on the aforementioned public datasets.
However this is due to the presence of a large number of similar and standard queries in the training and test set.
To reflect the performance in realistic settings, where users can input non-standard paraphrases of the queries, we propose to create more difficult subsets of the provided test sets to mimic the real-world setting.

Following \citet{Arora2020HINT3RT}, we create a subset of each test set with semantically dissimilar sentences from the training set. Instead of using ELMo \cite{peters-etal-2018-deep} and entailment scores, we use 
TF/IDF cosine distance to pick the most difficult examples from the original test sets. Each intent is treated separately during the selection process. First, all training utterances in a specific intent are tokenized (using simple white-space based tokenizer, ignoring punctuation). These tokenized training utterances are concatenated and transformed to TF/IDF vector space. Then, each testing example of the intent is transformed using the initialized TF/IDF transformer and cosine similarities with the transformed training set are calculated. Finally, $5$ least similar examples per intent are selected for inclusion to the difficult test set. For example, the CLINC150 dataset has $150$ intents, so our algorithm creates a test set of $750$ examples. Analogous process is used for the other two datasets. 
\footnote{We release the difficult subsets at \url{https://github.com/haodeqi/BenchmarkingIntentDetection} to facilitate repeatability and future research.}

\section{Experiment I: Comparison with Pretrained LMs}
\label{sec:pretrained-lm}

Pretrained LMs finetuned for intent detection have been shown to perform very well in recent literature, such as \cite{Casanueva+20}. Users can modify and adapt pretrained LMs to serve them as part of a scalable solution.
However, this often requires a complex solution design, an example of which can be found in \citet{yu2020financial}. 
In this work we evaluate and compare the commercial services with the following pretrained LMs: \textbf{USE$_{\text{base}}$}, i.e.,  Universal Sentence Encoder \cite{cer2018universal}; \textbf{Distilbert$_{\text{base}}$} \cite{sanh2020distilbert};
\textbf{BERT$_{\text{base}}$}, \textbf{BERT$_{\text{large}}$} \cite{Devlin+18}; and \textbf{RoBERTa$_{\text{base}}$} \cite{Liu+19}.


We compare Watson Assistant, RASA, and the aforementioned pretrained-LMs on the datasets and settings described in Section~\ref{datasets}, and measure the training time as well as accuracy.
\vspace{0.05in}
\par \noindent \textbf{Watson Assistant}
We evaluate both the \emph{classic} version of IBM Watson Assistant (WA) \footnote{\url{https://www.ibm.com/cloud/watson-assistant}} and the \mbox{\emph{enhanced}} version with improved intent detection algorithm.
Public API is used to train and evaluate the model. For training time, we measure the round-trip latency from sending the training request until we receive the status that the model is trained and available for serving. \footnote{Note that the training times may vary depending on the load on the web API.}
\vspace{0.05in}
\par \noindent \textbf{RASA}\footnote{\url{https://rasa.com}} The tool offers the flexibility to incorporate other open-source models such as Transformer-based \cite{Vaswani+17} models into the pipeline. For our experiments, we use the default training setting that trains a count-based feature ensemble with the DIETClassifier \cite{bunk2020diet}.
\vspace{0.05in}
\par \noindent \textbf{Pretrained LMs} For BERT-based models, we add a softmax classifier on top of the [CLS] token and finetune all layers. We use AdamW \cite{loshchilov2018decoupled} with $0.01$ weight decay and a linear learning rate scheduler. We choose a batch size of $32$, max sequence length $128$ and learning rate warmup for the first $10\%$ of the total iterations, peaking at $0.00004$. For training set variants of $5/30/$all examples per intent, we train for $50/18/5$ epochs, respectively. For USE$_{\text{base}}$ model, we train a softmax layer on top of the sentence representation and finetune all layers for 100 epochs. A learning rate of $0.05$ and batch size of $32$ are used for all training set variants. All models are trained with a single CPU core and a single K80 GPU.

\subsection{Results and Analysis}

\paragraph{Results in the full-set setting}

Table \ref{tab:opensource-accuracy-full} shows results of Watson Assistant, RASA and pretrained LMs on CLINC150, HWU64, and BANKING77. We train on the full training sets and report result on the full test sets, measured by accuracy. The overall best finetuned LM RoBERTa$_{\text{base}}$ achieves 1.5\% 
higher accuracy than Watson Assistant \emph{enhanced}. However, the improvement from finetuning large pretrained LMs requires more computational resources.

\begin{table}[!t]
\centering
\setlength{\tabcolsep}{0.2em}\resizebox{\linewidth}{!} {%
\begin{tabular}{c >{\centering\arraybackslash}p{1.5cm} >{\centering\arraybackslash}p{1.5cm}  >{\centering\arraybackslash}p{1.5cm}
>{\centering\arraybackslash}p{1.5cm}}
\toprule
             & \multicolumn{1}{c}{\underline{\textbf{CLINC150}}}           & \multicolumn{1}{c}{\underline{\textbf{HWU64}}}                & \multicolumn{1}{c}{\underline{\textbf{BANKING77}}}       &
             \multicolumn{1}{c}{\underline{\textbf{Average}}}\\
\toprule
WA classic           & 93.9       & 88.8      & 90.6    & 91.1                                 \\
WA enhanced           & 95.7 &  90.5 &  92.6  & 92.9\\
\midrule
RASA             & 89.4     & 84.9      & 89.9                        & 88.1      \\
Distilbert-base  & 96.3     & 91.7      & 92.1                        & 93.4      \\
BERT-base        & 96.8     & 91.6      & 93.3                        & 93.9      \\
BERT-large       & \textbf{97.1}   & 91.9   & 93.7                 & 94.2      \\
USE-base         & 94.7     & 88.9      & 89.9                        & 91.2     \\
RoBERTa-base     & 97.0      & \textbf{92.1}      & \textbf{94.1}  & \textbf{94.4} \\
\bottomrule              
\end{tabular}%

}
\vspace{-0.1in}
    \caption{\textbf{Accuracy on CLINC150, HWU64 and BANKING77 for Watson Assistant (WA), RASA and pretrained LMs}. Training is performed on the full train sets and evaluation on full test sets.}
    \label{tab:opensource-accuracy-full}
\end{table}
\begin{table}[!h]
\centering
\setlength{\tabcolsep}{0.2em}\resizebox{\linewidth}{!} {%
    \begin{tabular}{@{}ccccccc@{}}
     \toprule
     & \multicolumn{6}{c}{\underline{\textbf{CLINC150}}}                                                        \\
     \midrule
     &
      \multicolumn{2}{c}{\textbf{5 ex/class}} &
      \multicolumn{2}{c}{\textbf{30 ex/class}} &
      \multicolumn{2}{c}{\textbf{full}} \\
    \hline
     &
      Training   time &
      Accuracy &
      Training   time &
      Accuracy &
      Training   time &
      Accuracy \\
      \hline
    WA classic           & 0.58  & 78.1 & 0.78  & 90.3 & 1.04  & 93.9   \\
    WA enhanced          & 0.66  & 83.6 & 0.63  & 92.5 & 1.81  & 95.7   \\
    RASA                 & 1.25  & 53.2 & 5.6   & 79.4  & 13.93 & 89.4  \\
    Distilbert-base      & 15.23 & 82.2  & 31.65 & 93.2 & 35.98 & 96.3  \\
    BERT-base            & 29.67 & 83.8 & 61.43 & 94.7 & 71.08 & 96.8   \\
    BERT-large           & 125   & 87    & 280   & 95.8  & 270   & 97.1 \\
    USE-base             & 1.63  & 83.9 & 6.5   & 92.9  & 14.73 & 94.7  \\
    RoBERTa-base         & 33    & 86.3  & 85    & 95.4  & 90    & 97.0 \\
     \toprule
                 & \multicolumn{6}{c}{\underline{\textbf{HWU64}}}                                                                         \\
     \midrule
     &
      \multicolumn{2}{c}{\textbf{5 ex/class}} &
      \multicolumn{2}{c}{\textbf{30 ex/class}} &
      \multicolumn{2}{c}{\textbf{full}} \\
     \hline
     &
      Training   time &
      Accuracy &
      Training   time &
      Accuracy &
      Training   time &
      Accuracy \\
    \hline 
    WA classic          & 0.39  & 64.1 & 0.59  & 81.4  & 0.85  & 88.8 \\
    WA enhanced         & 0.75  & 71.0 & 0.54  & 86.2  & 0.82  & 90.5 \\
    RASA                & 0.67  & 43.7 & 2.17  & 72.4  & 9.43  & 84.9 \\
    Distilbert-base     & 6.32  & 71.1 & 13.92 & 86.3  & 20.35 & 91.7 \\
    BERT-base           & 12.73 & 70.1 & 27.18 & 87.5  & 39.48 & 91.6 \\
    BERT-large          & 52    & 77.3  & 120  & 89.3  & 175   & 91.9 \\
    USE-base            & 1.2   & 72.5  & 2.46  & 86.3  & 8.92  & 88.9  \\
    RoBERTa-base        & 13    & 71.7  & 40    & 88.8   & 60    & 92.1 \\
     \toprule
                 & \multicolumn{6}{c}{\underline{\textbf{BANKING77}}}                                                                     \\
     \midrule
     &
      \multicolumn{2}{c}{\textbf{5 ex/class}} &
      \multicolumn{2}{c}{\textbf{30 ex/class}} &
      \multicolumn{2}{c}{full} \\
    \hline
     &
      Training   time &
      Accuracy &
      Training   time &
      Accuracy &
      Training   time &
      Accuracy \\
    \hline
    WA classic       & 0.38  & 64.2 & 0.49  & 84.7 & 0.64  & 90.6 \\
    WA enhanced      & 0.65  & 69.9 & 0.52  & 87.0 & 1.22  & 92.6 \\
    RASA             & 0.89  & 45.1 & 3.67  & 81.6 & 15.45 & 89.9 \\
    Distilbert-base  & 7.87  & 69.8  & 16.83 & 87.8 & 20.35 & 92.1 \\
    BERT-base        & 15.23 & 68.3 & 32.72 & 88.9 & 38.75 & 93.3  \\
    BERT-large       & 92    & 71.2  & 210   & 89.9  & 175   & 93.7 \\
    USE-base         & 1.33  & 65.3 & 2.95  & 86.8 & 9.47  & 89.9  \\
    RoBERTa-base     & 17    & 75.9  & 42    & 90.4  & 57    & 94.1  \\
    \bottomrule
    \end{tabular}%
    }
    \caption{\textbf{Accuracy and training time (in minutes) comparing Watson Assistant (WA) with RASA and pretrained LMs}. We use $5/30/$all examples per intent on CLINC150, HWU64 and BANKING77 datasets. Results are the on the respective full test sets.}
    \label{tab:opensource-full}
\end{table}

\paragraph{Results in the few-shot setting}



Table~\ref{tab:opensource-full} shows results on few-shot setting for 5/30/all examples per intent on CLINC150, HWU64 and BANKING77 datasets on the full test sets. For experimental settings and dataset details, refer to Section~\ref{sec:pretrained-lm}.

\paragraph{Results in the difficult test setting}

\begin{table}[!t]
\centering
\setlength{\tabcolsep}{0.2em}\resizebox{\linewidth}{!} {%
\begin{tabular}{c >{\centering\arraybackslash}p{1.5cm} >{\centering\arraybackslash}p{1.5cm}  >{\centering\arraybackslash}p{1.5cm}}
\toprule
             & \multicolumn{1}{c}{\underline{\textbf{CLINC150}}}           & \multicolumn{1}{c}{\underline{\textbf{HWU64}}}                & \multicolumn{1}{c}{\underline{\textbf{BANKING77}}}            \\
\toprule
WA classic           & 79.3      & 83.4      & 75.2                               \\
WA enhanced           & 86.0  & 85.8  & 80.6 \\
\midrule
RASA             & 68.3      & 78.9      & 76.9                               \\
Distilbert-base  & 85.7      & 87.4      & 79.2                               \\
BERT-base        & 87.6     & 87.6      & 81.7                               \\
BERT-large       & \textbf{89.5}  &\textbf{89.2}  & \textbf{83.9}   \\
USE-base          & 81.6      & 83.4      & 74.5                               \\
RoBERTa-base     & 88.4      & 88.5      & 83.8           \\
\bottomrule              
\end{tabular}%

}

    \caption{\textbf{Accuracy on CLINC150, HWU64 and BANKING77 for Watson Assistant (WA), RASA and pretrained LMs}. Models are trained on full train sets and evaluated on Tfidf-difficult test sets.}
    \label{tab:opensource-accuracy-difficult}
\end{table}
\begin{table}[!h]
\centering
\setlength{\tabcolsep}{0.2em}\resizebox{\linewidth}{!} {%
    \begin{tabular}{@{}ccccccc@{}}
     \toprule
     & \multicolumn{6}{c}{\underline{\textbf{CLINC150}}}                                                        \\
     \midrule
     &
      \multicolumn{2}{c}{\textbf{5 ex/class}} &
      \multicolumn{2}{c}{\textbf{30 ex/class}} &
      \multicolumn{2}{c}{full} \\
    \hline
     &
      Training   time &
      Accuracy &
      Training   time &
      Accuracy &
      Training   time &
      Accuracy \\
      \hline
    WA classic            & 0.58  & 54.0    & 0.78  & 69.9 & 1.04  & 79.3  \\
    WA enhanced           & 0.66  & 65.1 & 0.63  & 76.7 & 1.81  & 86.0     \\
    RASA                  & 1.25  & 29.6  & 5.6   & 52.5 & 13.93 & 68.3  \\
    Distilbert-base       & 15.23 & 63.4  & 31.65 & 76.8  & 35.98 & 85.7  \\
    BERT-base             & 29.67 & 64.6  & 61.43 & 81.1 & 71.08 & 87.6   \\
    BERT-large            & 125   & 72.0    & 280   & 85.6  & 270   & 89.5   \\
    USE-base              & 1.63  & 66.6  & 6.5   & 77.5 & 14.73 & 81.6   \\
    RoBERTa-base          & 33    & 70.8  & 85    & 83.7  & 90    & 88.4   \\
     \toprule
                 & \multicolumn{6}{c}{\underline{\textbf{HWU64}}}                                                                         \\
     \midrule
     &
      \multicolumn{2}{c}{\textbf{5 ex/class}} &
      \multicolumn{2}{c}{\textbf{30 ex/class}} &
      \multicolumn{2}{c}{full} \\
     \hline
     &
      Training   time &
      Accuracy &
      Training   time &
      Accuracy &
      Training   time &
      Accuracy \\
    \hline 
    WA classic          & 0.39  & 53.9 & 0.59  & 72.3 & 0.85  & 83.4  \\
    WA enhanced         & 0.75  & 62.7 & 0.54  & 80.0    & 0.82  & 85.8   \\
    RASA                & 0.67  & 34.5 & 2.17  & 63.5 & 9.43  & 78.9  \\
    Distilbert-base     & 6.32  & 63.4 & 13.92 & 79.7 & 20.35 & 87.4  \\
    BERT-base           & 12.73 & 61.6 & 27.18 & 82.1  & 39.48 & 87.6  \\
    BERT-large          & 52    & 71.1  & 120   & 85.3  & 175   & 89.2   \\
    USE-base            & 1.2   & 66.3 & 2.46  & 79.8  & 8.92  & 83.4  \\
    RoBERTa-base        & 13    & 64.5  & 40    & 83.9  & 60    & 88.5   \\
     \toprule
                 & \multicolumn{6}{c}{\underline{\textbf{BANKING77}}}                                                                     \\
     \midrule
     &
      \multicolumn{2}{c}{\textbf{5 ex/class}} &
      \multicolumn{2}{c}{\textbf{30 ex/class}} &
      \multicolumn{2}{c}{full} \\
    \hline
     &
      Training   time &
      Accuracy &
      Training   time &
      Accuracy &
      Training   time &
      Accuracy \\
    \hline
    WA classic       & 0.38  & 43.2 & 0.49  & 64.5 & 0.64  & 75.2  \\
    WA enhanced      & 0.65  & 49.1 & 0.52  & 69.7 & 1.22  & 80.6  \\
    RASA             & 0.89  & 26.9 & 3.67  & 57.9 & 15.45 & 76.9  \\
    Distilbert-base  & 7.87  & 50.0    & 16.83 & 69.0 & 20.35 & 79.2  \\
    BERT-base        & 15.23 & 48.3 & 32.72 & 73.4 & 38.75 & 81.7  \\
    BERT-large       & 92    & 52.6  & 210   & 75.8  & 175   & 83.9   \\
    USE-base         & 1.33  & 44.5 & 2.95  & 68.6 & 9.47  & 74.5  \\
    RoBERTa-base     & 17    & 57.1  & 42    & 75.7  & 57    & 83.8   \\
    \bottomrule
    \end{tabular}%
    }
    \caption{\textbf{Accuracy and training time (in minutes) comparing Watson Assistant (WA) with RASA and pretrained LMs}. We use $5/30/$all examples per intent on CLINC150, HWU64 and BANKING77 datasets. Results are the on the respective Tfidf-difficult test sets.}
    \label{tab:opensource-difficult}
\end{table}

Table ~\ref{tab:opensource-accuracy-difficult} shows results on our difficult test sets. We observe that there is a significant drop in accuracy compared to the full test set, going from $90\%$+ to $80\%$s. This shows that these test sets are indeed more difficult for all algorithms, and they provide a better testbed for identifying the robustness of a intent detection system. In addition, we conduct the comparison in few-shot settings, where we use $5$ examples per intent for training, and increase to $30$ and full training sets. The complete set of results of few-shot setting on the difficult test sets can be found in Table \ref{tab:opensource-difficult}. Results show that BERT$_{\text{large}}$ performs the best in terms of accuracy. However, Watson Assistant still stands on top considering the trade-off between training time and accuracy. 


\begin{figure}[t]
\includegraphics[width=\linewidth]{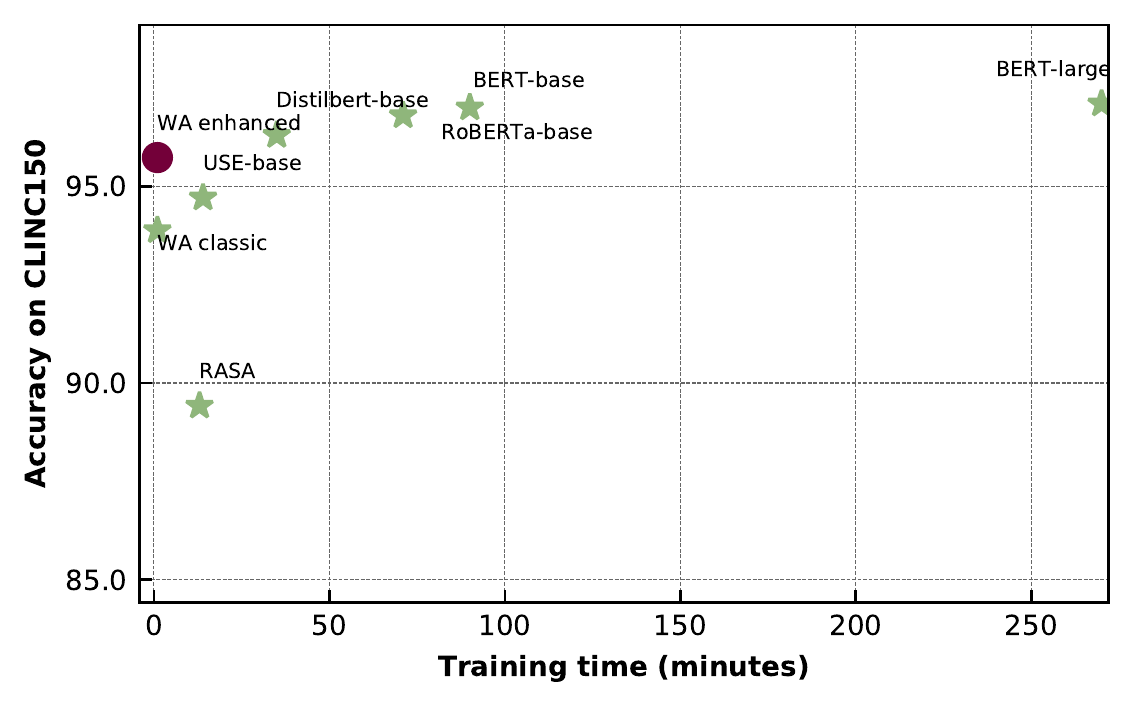}

\caption{\textbf{Training time vs. accuracy on CLINC150} for Watson Assistant (WA), RASA and pretrained LMs. Full training set and test set are used. All methods except Watson Assistant are trained using GPU. Watson Assistant offers the best trade-off between training time and accuracy. }
\label{scatter-clinc}
\end{figure}

\paragraph{Training time vs accuracy trade-off}
We report the training times and resources used for all models across the three datasets in Table~\ref{tab:opensource-time}. We observe that the pretrained LMs require significantly more training time compared to Watson Assistant. For example, RoBERTa$_{\text{base}}$ achieves comparable performance to Watson Assistant but requires 90 minutes training time on CLINC150. Figure~\ref{scatter-clinc} shows a visualization of accuracy and training time for each model.
Watson Assistant offers the best trade-off in terms of accuracy vs. training time.

\begin{table}[!t]
\centering
\setlength{\tabcolsep}{0.2em}\resizebox{\linewidth}{!} {%
\begin{tabular}{ccccc}
\toprule
\multirow{2}{*}{\textbf{Algorithm}} & \multirow{2}{*}{\textbf{Resources}} & \textbf{CLINC150}      & \textbf{HWU}           & \textbf{BANKING77}     \\
                           &          &  Training time &  Training time & Training time \\
\toprule
WA classic                     & -      & \textbf{1.04}          & 0.85          & \textbf{0.64}          \\
WA enhanced                     & -      & 1.81          & \textbf{0.82}          & 1.22          \\
\hline
RASA                       & GPU      & 13.93         & 9.43          & 15.45         \\
Distilbert-base            & GPU      & 35.98         & 20.35         & 20.35         \\
BERT-base                  & GPU      & 71.08         & 39.48         & 38.75         \\
BERT-large                 & GPU      & 270           & 175           & 175           \\
USE-base                   & GPU      & 14.73         & 8.92          & 9.47          \\
RoBERTa-base               & GPU      & 90            & 60            & 57            \\
\bottomrule
\end{tabular}%
}
    \caption{\textbf{Training time (in minutes) and resource requirements} for Watson Assistant (WA), RASA and pretrained LMs. Training is performed on full training sets. All methods except for Watson Assistant are trained using a single NVIDIA K80 GPU.}
    \label{tab:opensource-time}
\end{table}

We report results on HINT3 datasets for completeness and are discussed in Section \ref{Experiments2} Table \ref{tab:market-hint3-avg}.

\section{Experimental II: Comparison among Commercial Solutions}

\label{Experiments2}

Finally, we conduct comparison studies among commercial services. Commercial solutions are more suitable for enterprise customers and are designed for users who have limited knowledge of machine learning and natural language processing. One of the challenges in comparing the performance of commercial services and designing experiments lies in the fact that most service providers have terms of use prohibiting any type of benchmarking on their services. To overcome this challenge, we use the prior benchmarking study from \citet{Arora2020HINT3RT} to obtain the performance of existing commercial solutions. In this benchmark, HINT3 dataset collection is used which contains three tasks with small amounts of training data. We extended the study by including the results on the Watson Assistant service.

In this section, we evaluate the performance of the following commercial solutions: \mbox{\textbf{IBM Watson Assistant}}\footnote{\url{https://www.ibm.com/cloud/watson-assistant}}, \textbf{Google Dialogflow}\footnote{\url{https://cloud.google.com/dialogflow}}, \textbf{Microsoft LUIS}\footnote{\url{https://www.luis.ai}}, and the open-source solution \textbf{RASA}\footnote{\url{https://rasa.com}}.
We use the prior benchmarking study from \citet{Arora2020HINT3RT} to obtain the performance of these commercial solutions, except for Watson Assistant. 

\subsection{Datasets}

\textbf{HINT3} is a collection of three datasets: SOFMattress, Curekart, and Powerplay11. The statistics of the datasets are shown in Table \ref{tab:hint3-stats}. Each dataset has two training set variants referred to as \emph{full} and \emph{subset}. The subset variant was created by discarding semantically similar sentences using ELMo \cite{peters-etal-2018-deep} and entailment score > $0.6$ \cite{Arora2020HINT3RT}. We used both variants of the training data in our experiments. The test sets contain both in-scope and out-of-scope examples. 

\begin{table}[!t]
    \centering
    \setlength{\tabcolsep}{0.2em}\resizebox{\linewidth}{!} {%
    \begin{tabular}{c >{\centering\arraybackslash}p{1.5cm} >{\centering\arraybackslash}p{2.3cm} >{\centering\arraybackslash}p{1.5cm} >{\centering\arraybackslash}p{2.3cm} >{\centering\arraybackslash}p{1.5cm} >{\centering\arraybackslash}p{2.3cm}}
    \toprule
    & \multicolumn{2}{c}{\underline{\textbf{SOFMattress}}} & \multicolumn{2}{c}{\underline{\textbf{Curekart}}} & \multicolumn{2}{c}{\underline{\textbf{Powerplay11}}} \\
     & Train & Test (in-scope / out-of-scope) & Train & Test (in-scope / out-of-scope) & Train & Test (in-scope / out-of-scope) \\
    \toprule
Full                     & 328             & 231/166           & 600           & 452/539          & 471             & 275/708           \\
Subset                   & 180             & 231/166           & 413           & 452/539          & 261             & 275/708          \\
\bottomrule
\end{tabular}%
}
\caption{\textbf{HINT3 training and test set statistics}. HINT3 consists of three datasets - SOFMattress, Curekart and Powerplay11. Each training set contains two versions - \emph{Full} and \emph{Subset}. The test set is also broken down into in-scope queries and out-of-scope queries.}
\label{tab:hint3-stats}
\end{table}

\subsection{Experimental Setup}
We use the same experimental setup as described in \citet{Arora2020HINT3RT}. Following their methodology, we use a confidence threshold of $0.1$. For the BERT model reported in their paper, they used BERT$_{\text{base}}$ and finetuned all layers upto 50 epochs, learning rate of $4\times10^{-5}$ with warmup period of 0.1 and early stopping.


\subsection{Results}
Table \ref{tab:market-hint3} shows full results on the in-scope test \mbox{examples} of each dataset measured by accuracy using a confidence threshold of $0.1$. 

On average across the datasets (Table \ref{tab:market-hint3-avg}), Watson Assistant \emph{enhanced} achieves $73.8\%$ accuracy when trained on the full training sets and evaluated on the in-scope examples, outperforming DialogFlow by $4.57\%$, and LUIS by $13.87\%$. Training on the subset variant of the datasets, Watson Assistant also consistently outperforms the other commercial solutions. It is worth noting that Watson Assistant also does better than BERT by $4.4\%$ on average.

\begin{table}[!t]
    \centering
    \small
    \setlength{\tabcolsep}{0.2em}\resizebox{\linewidth}{!} {%
    \begin{tabular}{l|cccccc}
    \toprule
    & \multicolumn{2}{c}{\underline{\textbf{SOFMattress}}} & \multicolumn{2}{c}{\underline{\textbf{Curekart}}} & \multicolumn{2}{c}{\underline{\textbf{Powerplay11}}} \\
     & full & subset & full & subset & full & subset \\
    \toprule
    WA classic & 73.6 & 66.2 & 83.2 & 79.9 & 63.3 & 57.1 \\
    WA enhanced & \textbf{74.0} & \textbf{68.4} & \textbf{86.7} & \textbf{85.4} & 60.7 & 57.8 \\
    \hline
    Dialogflow & 73.1 & 65.3 & 75.0 & 71.2 & 59.6 & 55.6 \\
    RASA & 69.2 & 56.2 & 84.0 & 80.5 & 49.0 & 38.5 \\
    LUIS & 59.3 & 49.3 & 72.5 & 71.6 & 48.0 & 44.0 \\
    Haptik & 72.2 & 64.0 & 80.3 & 79.8 & \textbf{66.5} & \textbf{59.2} \\
    BERT  & 73.5 & 57.1 & 83.6 & 82.3 & 58.5 & 53.0 \\
    \bottomrule
    \end{tabular}%
    }
    \vspace{-0.1in}
    \caption{\textbf{In-scope Accuracy on HINT3 using commercial solutions.} We report the in-scope accuracy with a threshold of 0.1 for various intent detection methods. Results for all methods except Watson Assistant (WA) are obtained from \cite{Arora2020HINT3RT}.}
    \label{tab:market-hint3}
\end{table}

\begin{table}[!t]
    \centering
    \small
    \setlength{\tabcolsep}{1em}\resizebox{\linewidth}{!} {%
    \begin{tabular}{l|ccc}
    \toprule
    & {\underline{\textbf{Full}}} & {\underline{\textbf{Subset}}} & {\underline{\textbf{Average}}} \\
     
    \toprule
    WA classic & 73.4 & 67.7 & 70.6 \\
    WA enhanced & \textbf{73.8} & \textbf{70.5} & \textbf{72.2}\\
    \hline
    Dialogflow & 69.2  & 64.0 & 66.6\\
    RASA & 67.4 & 58.4 & 62.9 \\
    LUIS &  59.9 & 54.6 & 57.5\\
    Haptik &73.0 & 67.6 & 70.3\\
    BERT & 71.9 & 64.1 & 68.0\\
    \bottomrule
    \end{tabular}%
    }
    \vspace{-0.1in}
    \caption{\textbf{Average In-scope Accuracy on HINT3 using commercial solutions.}
    We report the average in-scope accuracy across the three datasets with a threshold of 0.1 on Full and Subset versions of the HINT3 collection. Results for all methods except Watson Assistant (WA) are obtained from \cite{Arora2020HINT3RT}.}
    \label{tab:market-hint3-avg}
\end{table}

 \section{Conclusion}

 We proposed a new methodology to assess the performance of intent detection "in the wild" in task-oriented dialog systems. In practice, the platforms developed for building and deploying virtual assistants have to consider several scenarios and trade-offs. These systems have to train the best performing models in few-shot settings, strike a compromise between training time and accuracy, and adapt seamlessly to a wide range of domains.

We compare the performance of leading commercial services which are designed to develop task-oriented dialog systems on the publicly available datasets and also compared their performance against popular pretrained LMs. Our results demonstrate that Watson Assistant outperforms market competitors on the HINT3 dataset collection, which comprises real-world queries. Our results also show that Watson Assistant is competitive with pretrained LMs across a wide range of datasets and settings but trains much faster - which is a key factor in usability of a commercial conversational AI solution.

\bibliography{anthology,custom}
\bibliographystyle{acl_natbib}

\clearpage

\end{document}